\documentclass[letterpaper, 10 pt, conference]{ieeeconf}  

\IEEEoverridecommandlockouts                              

\usepackage{graphics} 
\usepackage{epsfig} 

\usepackage{booktabs} 

\title{\LARGE \bf
A Multi-task Learning Framework for Grasping-Position Detection and Few-Shot Classification
}

\author{Yasuto Yokota, Kanata Suzuki, Yuzi Kanazawa,  and Tomoyoshi Takebayashi
\thanks{All authors are with Fujitsu Laboratories LTD., Kanagawa 211-8588, Japan.
        {\tt\small yokota.yasuto@fujitsu.com}}%
}

\begin{document}

\maketitle
\thispagestyle{empty}
\pagestyle{empty}

\begin{abstract}
It is a big problem that a model of deep learning for a picking robot needs many labeled images.
Operating costs of retraining a model becomes very expensive because the object shape of a product or a part often is changed in a factory. 
It is important to reduce the amount of labeled images required to train a model for a picking robot.
In this study, we propose a multi-task learning framework for few-shot classification using feature vectors from an intermediate layer of a model that detects grasping positions.
In the field of manufacturing, multi-task for shape classification and grasping-position detection is often required for picking robots.
Prior multi-task learning studies include methods to learn one task with feature vectors from a deep neural network (DNN) learned for another task.
However, the DNN that was used to detect grasping positions has two problems with respect to extracting feature vectors from a layer for shape classification:
(1) Because each layer of the grasping position detection DNN is activated by all objects in the input image, it is necessary to refine the features for each grasping position.
(2) It is necessary to select a layer to extract the features suitable for shape classification.
To tackle these issues, we propose a method to refine the features for each grasping position and to select features from the optimal layer of the DNN.
We then evaluated the shape classification accuracy using these features from the grasping positions.
Our results confirm that our proposed framework can classify object shapes even when the input image includes multiple objects and the number of images available for training is small.
\end{abstract}

\section{INTRODUCTION}
In the field of manufacturing, picking robots often require multi-task, specifically shape classification and grasping-position detection, to pick a object.
For these tasks, conventionally, large numbers of parameters needed to be manually adjusted for each robot.
Therefore, in recent years, many studies have applied deep neural networks (DNNs), which have multiple layers to implement these tasks\cite{r1}\cite{r2}\cite{r3}\cite{r4} and have achieved high accuracy.
\par
DNN automatically learns the extraction of the features needed to solve tasks from a large amount of data.
In related tasks such as object detection and classification, multi-task learning, which involves learning multiple tasks simultaneously, learns common features of tasks and improves the prediction accuracy compared with the accuracy of learning each task individually.
If the settings (e.g., the number of predicted targets) of all the tasks are similar, a single DNN is used for the learning[5], and it can be end-to-end trained.
However the object shape of a product or a pert often change in a factory.
It is difficult to make many new labeled images and to retrain the model, because of large costs.
\par
\begin{figure}[t]
\begin{center}
\includegraphics[width=8.0cm]{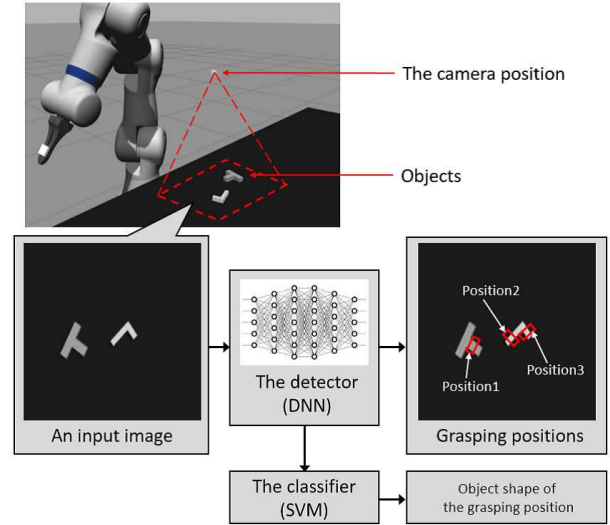}
\caption{
The workflow for the grasping position detection and shape classification.
A camera is used to photograph one or several objects placed on a workbench to create an input image.
The DNN predicts multiple grasping positions for the objects in the input image.
The SVM classifies the shapes of the objects for each grasping position.
}
\label{fig:picking task}
\end{center}
\end{figure}

Otherwise, multiple learners such as DNN or support vector machine (SVM) are generally used\cite{r6}\cite{r7}\cite{r8} because this simplifies the DNN design and creation of training data.
In this study, using two learners (Fig.\ref{fig:picking task}), we perform one task to predict multiple grasping positions from an input image and another task to classify the shape of the object using the features for each grasping position.
The DNN which detects grasping positions doesn't have to be retrained, when the object shape is changed.
Only the SVM which classifies object shape need to be retrained using a small amount of labeled images.
Our framework can reduce costs to make new labeled images.
\par

Transfer learning such as few-shot learning\cite{r6}\cite{r9} is often used when multiple tasks are related.
Using the features of the DNN that has learned one task, the other task can be learned efficiently.
It is usual to apply transfer learning to similar tasks\cite{r10}\cite{r11}; however, we consider this method to also be applicable to the tasks in this study.
Because object shape is an important factor in grasping-position detection, it is conjectured that features related to the object shape appear in an intermediate layer of the DNN learned for detecting the grasping positions.
For this reason, the shape classification of an object in an input image should be possible using these features\cite{r12}. 
\par
However, the DNN used to detect the grasping positions has two problems with respect to extracting feature vectors from a layer.
(1) Because each layer of the DNN for grasping-position detection is activated by all the objects in the input image, it is necessary to refine the features for each grasping position.
(2) The layer to extract the features suitable for shape classification needs to be selected.
To attack the above problems, we propose a method to refine the features for each grasping position and to select the features from the optimal layer of the DNN.
\par
It is proposed to refine the features for each grasping position via a method used in feature visualization\cite{r13}\cite{r14}.
This method visualizes the feature that contributes to the output result via guided backpropagation, which calculates the gradients of each layer for a specific final output of the DNN.
In our method, a solution to problem (1) is to refine the features.
This requires the gradients, which are calculated for each node of the intermediate layers for every grasping position.
\par
For problem (2), one solution is to select the optimal intermediate layer.
Our method creates SVMs of every layer using the training data and compares the classification accuracy; then, it uses the SVM of the highest accuracy layer to classify the test data.
The layer with the highest accuracy in the training data can extract features suitable for shape classification; therefore, we expect to also achieve high accuracy for the test data.
Moreover, the calculation cost is reduced when using low-dimensional feature vectors; the activations of the intermediate layers have reduced dimensions due to the dimension reduction.
\par
In addition, our proposed method has the effect of streamlining the learning of the shape classification task.
We achieve few-shot learning for this task due to our use of the optimal intermediate features for the shape classification.
\par
The main contributions of this paper are as follows:
\begin{itemize}
\item The realization of few-shot learning for shape classification of an object whose grasping positions are detected; and
\item A new proposed framework to select features suitable for multi-task.
\end{itemize}

\par
The rest of the paper is organized as follows. Section 2 reviews related studies.
Section 3 introduces our approach in detail.
Section 4 explains the experimental setup used to evaluate the proposed method.
Section 5 discusses the experimental results.
Section 6 describes our conclusions and future directions of study.

\begin{figure*}[t]
\begin{center}
\includegraphics[width=17.0cm]{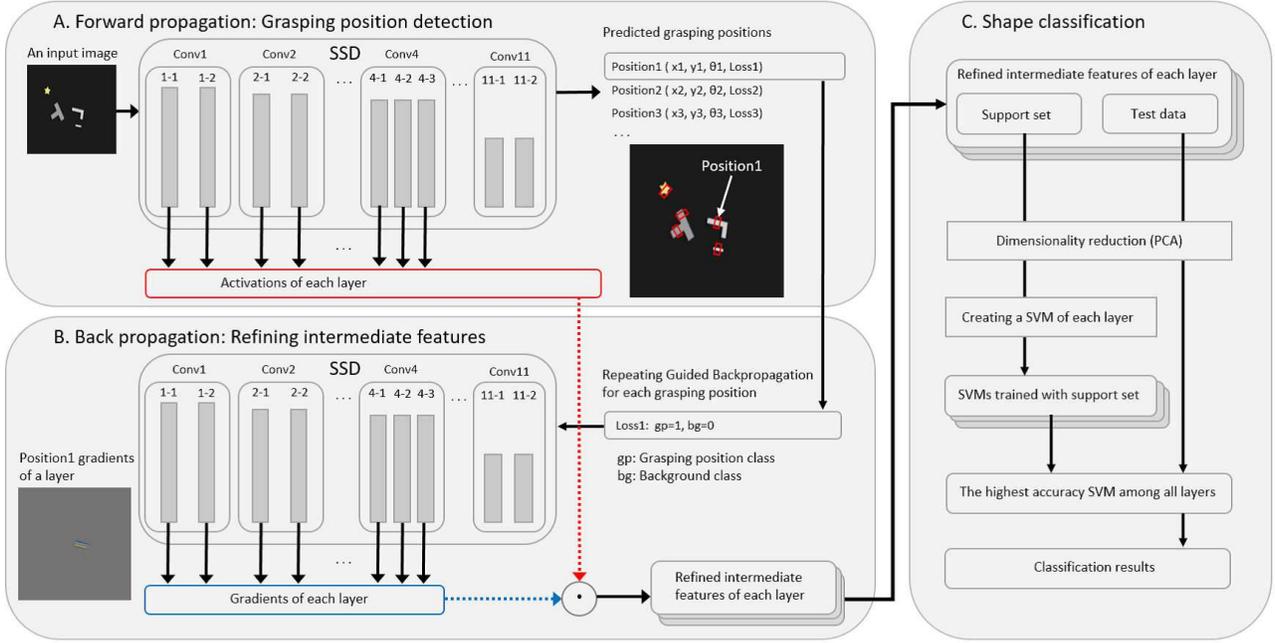}
\caption{
Overview of the proposed approach.
(A) The activations of each convolutional layer (from layer 1-1 to 4-3) at the time of grasping-position detection are stored.
(B) The intermediate features for each layer are refined by multiplying the activation by the gradient of the guided backpropagation.
This is repeated for each grasping position, and the intermediate features are extracted for each grasping position.
(C) The dimensions of the intermediate features are reduced.
The SVMs for each layer are created and trained with the features.
The SVM of the highest accuracy layer for the support set is used for the object shape classification of the test data.
}
\label{fig:overview}
\end{center}
\end{figure*}

\section{RELATED WORK}
In this section, we review related studies concerning detecting grasping positions and transfer learning.
Our approach employs shape classification using intermediate features in the DNN generated for detecting grasping positions.
However, few previous studies have focused on the use of intermediate features in DNNs that learn other tasks.

\subsection{Detecting grasping position}
As an example of previous studies applying DNN to the problem of grasping-position detection, Levine et al.\cite{r15} introduced a method to grasp objects of various shapes in an input image using deep reinforcement learning.
Active learning\cite{r2}\cite{r3} adds training data sequentially following the trial results.
Bousmalis et al.\cite{r1} proposed a method to improve the grasping accuracy using simulation images.
However, these studies focus on grasping-position detection and do not consider shape classification.
\par
Jang et al. proposed Grasp2Vec\cite{r16}, which learns grasping-position detection and shape classification simultaneously.
This makes it possible to grasp objects of a specified shape by embedding the grasping position and the object shape in the same feature space.
However, this method requires enormous, 400k or more, training data created by a real robot and it takes a large amount of time to learn.
\par
Our framework can perform grasping-position detection and shape classification simultaneously and can learn shape classification with less training data than conventional methods.

\subsection{Transfer learning}
Conventional studies on transfer learning have used a method to streamline learning that involves transferring the weight of a DNN pre-trained on a large data set, such as ImageNet, to another task\cite{r17}.
Zero-shot learning\cite{r9}\cite{r10}\cite{r11}\cite{r18} is an approach that can be used to classify unknown classes.
Because these methods can only be applied to the same task as the pre-trained task, they cannot be used for two different tasks, as in the case of this study.
\par
Conversely, one-shot learning\cite{r19}\cite{r20} and few-shot learning\cite{r6}\cite{r7}\cite{r21} can be used to solve different tasks.
Most such methods consist of two learners.
One learner solves the task to extract the features of the input data, and the other solves the task to classify the input data with the features.
By pre-training a DNN using large amounts of labeled data for one task, it is possible to solve the other task with a small amount of labeled data (called a support set).
\par
Suzuki et al.\cite{r22} and Kase et al.\cite{r23} have used transfer learning in the field of robotics.
In their studies, the features of the intermediate layer of the autoencoder for inputting the camera image were used by the other learner for the robot motion control.
However, in a grasping-position detection DNN, the intermediate layer in which the feature vectors of the object shape appears is not constant.
\par
Our proposed approach, which is an extension of few-shot learning, classifies the shape of the object using intermediate features extracted from the grasping-position detection DNN.

\section{METHOD}
An outline of the proposed framework is shown in Figure\ref{fig:overview}.
We used a DNN model with Single Shot Multibox Detector (SSD)\cite{r5}\cite{r24} to detect the grasping positions and support vector machine (SVM) to classify the object shapes.
First, the SSD model was trained only to detect the grasping positions\cite{r12}.
The intermediate layer activations of each layer were then refined via backpropagation for each detected grasping position to solve the shape classification task.
By extracting the activations, which contributed to the prediction, from the intermediate layer, it is possible to select features related to a specific object even in an input image in which objects of various shapes are mixed.
Next, dimension reduction was performed to extract features suitable for classification from the above interlayer activations.
Comparing this to the features of the support set prepared using the same above procedure, the shape of an object could be classified.
\par
Subsection A describes the grasping-position detection method using SSD, Subsection B explains the details of the method to refine the interlayer activations, and Subsection C provides the details of the object shape classification method.

\subsection{Grasping-position detection using SSD}
Here, we introduce our method for the grasping-position detection\cite{r3}.
SSD is a popular algorithm for DNN in object detection and can detect the positions of multiple objects in an input image.
Numerous bounding boxes of various sizes are set by default.
Each bounding box outputs the coordinates of the detected objects.
The SSD model has 23 convolution layers from the 1-1 layer to the 11-2 layer.
\par
In this paper, we define detected objects as being in the grasping position class as opposed to the background class.
We extended the SSD model to be able to predict the grip angles.
Each bounding box is therefore trained to predict the grasping position rectangle $(x,y,h,w)$ and the grip angle $\theta$.
At the time of inference, a bounding box, which is determined to be in the grasping position class, outputs the grasping position and the grip angle.

\subsection{Refining intermediate activations using backpropagation}
In some cases, an input image has multiple objects and grasping positions.
Therefore, it is necessary to refine and select the features that contributed to the prediction results from the intermediate activations for each detected grasping position.
To measure the degree of contribution to the inference result, we use the magnitude of the gradient of each node of the intermediate layers.
The gradients are calculated via backpropagation with the loss of the correct class as 1 and the loss of the other classes as 0.
According to the above setting, the gradient of the loss function is considered to become large in a node that makes a large contribution to the output of the correct class.
\par

Our method defines only the grasping position class of the detected grasping position bounding box as the correct class and backpropagates only that class for each grasping position.
We use guided backpropagation\cite{r13}, which does not attenuate the output of each node during the backpropagation in order to calculate the gradient in layers close to the input with less noise.
In guided backpropagation, the backpropagation is performed using the activation function shown in Eq. (1).
Here, the value of an activation during forward propagation in node $i$ of layer $l$ is $f^{l}_{i}$ and the value of a gradient during backpropagation is $R^{l}_{i}$.

\begin{eqnarray}
R^{l}_{i} = (f^{l}_{i} > 0) \cdot (R^{l+1}_{i} > 0) \cdot R^{l+1}_{i}
\end{eqnarray}

\par
The intermediate activation $F^{l}_{i}$ of the node contributing to the detected grasping position is emphasized and is minimized otherwise.
Therefore, only the feature vectors of the target grasping position are assumed to appear in each intermediate layer.
We can extract the features that are refined for each detected grasping position from the intermediate layers.

\begin{eqnarray}
F^{l}_{i} = f^{l}_{i} \cdot R^{l}_{i}
\end{eqnarray}


\subsection{Shape classification using intermediate features}
The proposed method includes two steps: prework and training SVMs.

\subsubsection{Prework}
The SSD is trained for the grasping-position detection task using the training set, whose correct label is a rectangle indicating the grasping position.
After learning, the SSD predicts the grasping positions using the support set.
We extract and accumulate the refined intermediate activations of each grasping position and save the object shape information of the label to train the SVMs.

\subsubsection{Training SVMs}
Our method reduces the dimensions of the intermediate activations for each layer.
We create SVMs for the shape classification and train them with the features, which are the above compressed activations.
To select the optimal intermediate layer for the shape classification, we calculate the classification accuracies for all the SVMs using the features of the support set.
Here, the penalty parameter of the error automatically optimized for the support set by a grid search is used for each SVM.
Then, the SVM with the highest accuracy is used for the shape classification of the test data.

\par
We simultaneously reduce the dimensions of the test data and the support set.
This is done to retain the many features that are highly correlated with the object shape after dimension reduction.


\begin{figure*}[t]
\begin{center}
\includegraphics[width=17.0cm]{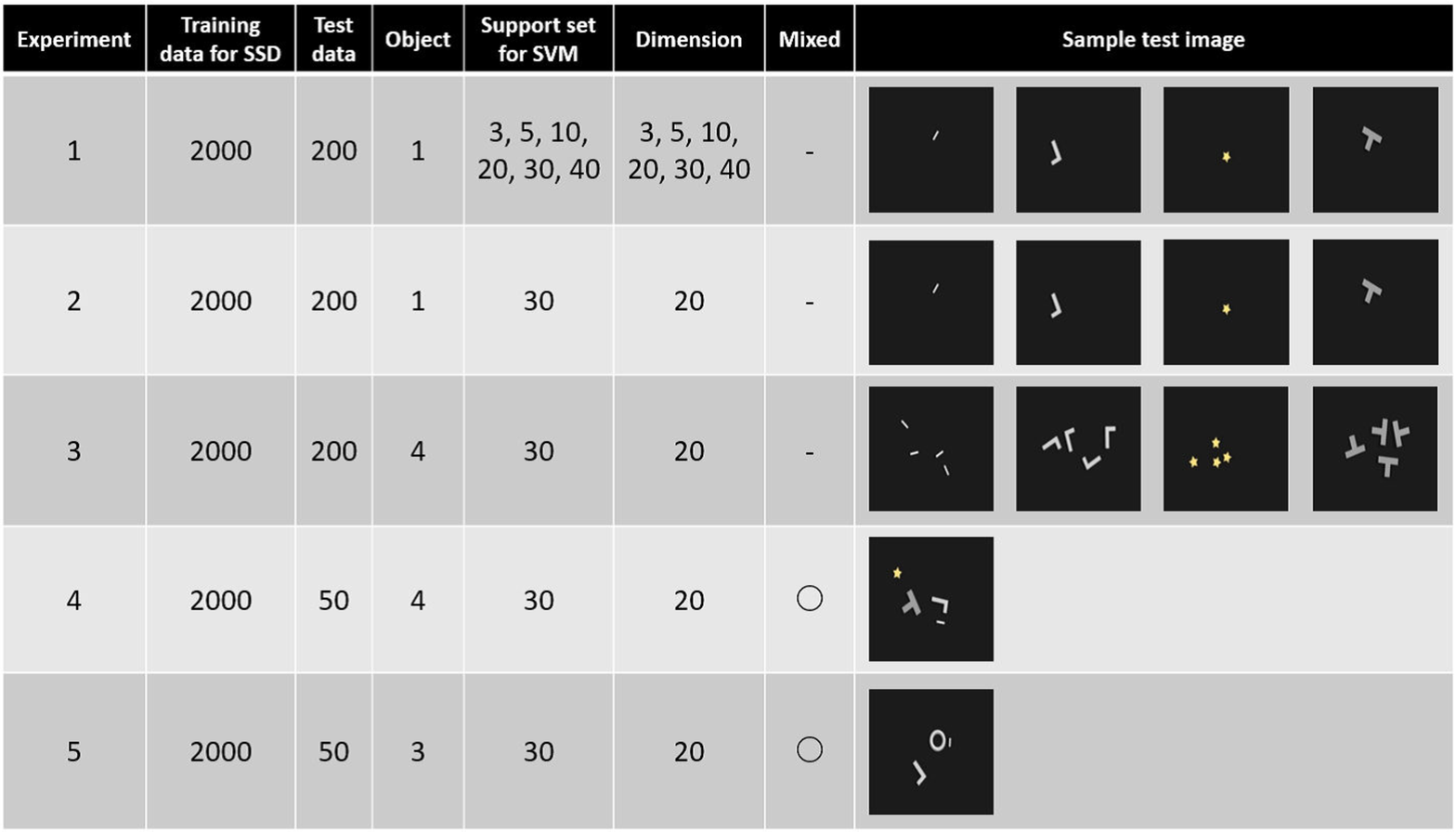}
\caption{
The setting of each experiment.
Training data for SSD represents the number of training data for the grasping-position detection;
Test data indicates the number of test data for the experimental evaluation; Object indicates the number of objects in the test data image; Support set for SVM indicates the number of support images used for the shape classification with SVM;
Dimension indicates the number of the principal component after the dimension reduction by PCA; and
Mixed is marked with a circle when objects of different shapes are placed in the test image.
}
\label{fig:sample_image}
\end{center}
\end{figure*}

\begin{figure}[t]
\begin{center}
\includegraphics[width=8.0cm]{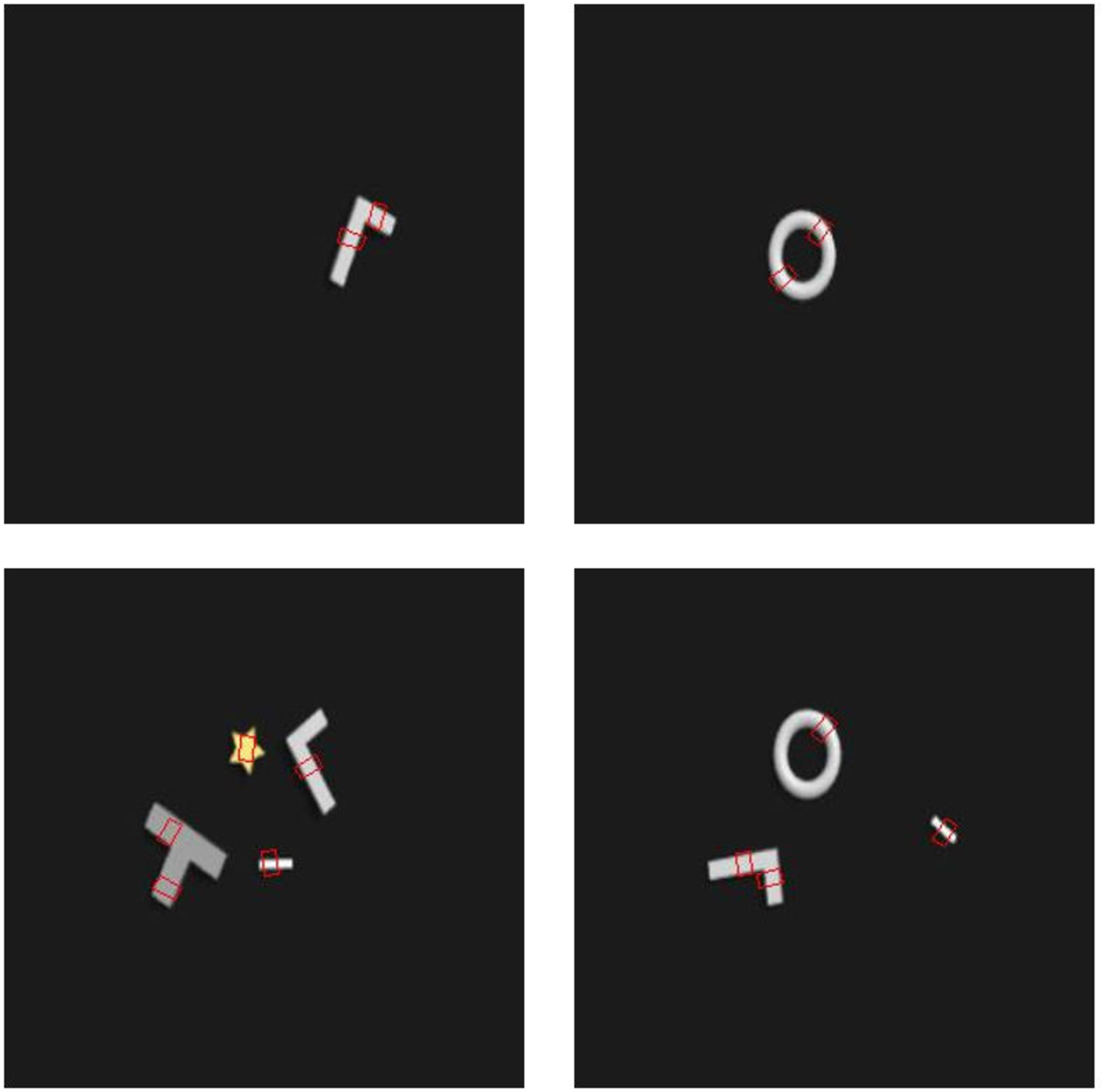}
\caption{
Prediction results of grasping position.
The training set for the SSD doesn't contain an image of a ring object.
The SSD can detect grasping positions of a ring without retraining.
}
\label{fig:prediction_results}
\end{center}
\end{figure}

\section{EXPERIMENTAL SETUP}
In this study, we propose a framework to classify the object shape using the intermediate features of SSD, which learn only the grasping-position detection task.
To evaluate the proposed approach, experiments were conducted with the following settings.

\subsection{Settings for learning}
In these experiments, the SSD outputs the grasping positions of objects in the input images (300$\times$300$\times$3[pix]).
We prepared 2000 sets of images, in which the labels were set at a plurality of positions for one object, as the training set for the SSD, and trained it in 200 epochs using Adam\cite{r25}.

\par
We used the images and the labels of the grasping position, which were automatically generated by Gazebo simulator\cite{r26}.
An image in the training set or the support set had any one of four types of shapes (cylinder, L-shape, star, and T-shape).
The objects were placed in random positions in the images.
Coordinates and angles of the grasping positions were set to positions suitable to grasp the object.
\par
We used principal component analysis (PCA) for the dimension reduction and RBF for the kernel of the SVMs.
Due to the structure of the SSD, some outputs did not pass through the higher layers (layer 5-1 and higher).
The target intermediate layers were the lower convolution layers (10 layers in total) from layer 1-1 to layer 4-3 of the SSD.

\subsection{Experimental settings}
To evaluate our approach, we conducted the following five experiments in terms of the number of images for support set (called support images), the number of dimensions and the classification accuracy.
The details of the experimental setup are shown in Figure\ref{fig:sample_image}.

\subsubsection{The number of support images and dimensions}
In Experiment 1, we confirmed the number of support images and dimensions required for the shape classification.
The number of support images in a support set used for few-shot learning\cite{r6}\cite{r7}\cite{r21} is generally in the range of several to several tens per class.
We set six cases (3, 5, 10, 20, 30, 40) for the number of support images per class and six cases (3, 5, 10, 20, 30, 40) for the number of dimensions.
The test data consisted of 50 images per shape (200 in total) in which one object was placed in a random position.

\subsubsection{Classification accuracy}
We measured the accuracy of the object shape classification for the detected grasping position.
Based on the results of Experiment 1, which are described later, the number of images per shape type in the support set was 30 and the number of dimensions after the dimension reduction with PCA was 20.
\par
Experiment 2 compared the accuracy of the support set to the accuracy of the test data for each layer and evaluated the automatic selection of the intermediate layer using the proposed method.
The test data consisted of 50 images per shape (200 in total) in which one object was placed in a random position.
\par
Experiment 3 evaluated the accuracy for an input image that had multiple objects of the same shape.
To confirm the effect of the proposed method for this case, we compared our results to the accuracy when we did not use our method.
In Experiment 3, the test data consisted of 50 images per shape (200 in total) in which four objects of the same shape were placed in random positions.
\par
Experiment 4 evaluated the accuracy for an input image that had multiple objects of different shapes.
To confirm the effect of the proposed method for this case, we compared our results to the accuracy when we did not use our method.
The test data consisted of 50 images in which four objects (one object for each shape) were placed in random positions.
\par
Experiment 5 evaluated the accuracy for an input image that contained a ring object which the SSD didn't learn.
The SSD can detect grasping positions of a ring(Fig.\ref{fig:prediction_results}), because the width of four objects which were learned is similar to the width of ring object.
The test data consisted of 50 images in which three objects (one was a ring object) were placed in random positions.

\section{RESULTS AND DISCUSSION}
\subsection{The number of support images and dimensions}
In Experiment 1, 226 grasping positions were detected for the 200 test data images.
The accuracy rate of the shape classification is shown in Table \ref{table:exp1}.
\par
The results show that increasing the support images and dimensions increases the accuracy.
When the number of support images and dimensions is 40, the accuracy rate is 1.000.
Even when the number of support images is 30 and the number of dimensions is 20, the accuracy rate is 0.97 or more, which is sufficient for shape classification.
These results confirm that it is possible to use a small amount of labeled data to classify the shape of an object whose grasping position was detected by the proposed framework.
However, this result uses images created by a simulator.
It is likely that more support images will be needed in a real environment because the images will become more complex.

\begin{table}[tb]

  \begin{center}
	\label{table:exp1}
	\begin{tabular}{|c|cccccc|}

		\multicolumn{7}{c}{TABLE I: Results of Experiment 1} \\

		\hline
		Dimensions & \multicolumn{6}{|c|}{Support images per shape type} \\
		 & 3 & 5 & 10 & 20 & 30 & 40 \\ \hline\hline
		3 & 0.566 & 0.544 & 0.730 & 0.721 & 0.788 & 0.796 \\
		5 & 0.637 & 0.593 & 0.739 & 0.748 & 0.885 & 0.894 \\
		10 & 0.664 & 0.668 & 0.841 & 0.850 & 0.960 & 0.987 \\
		20 & 0.655 & 0.699 & 0.788 & 0.916 & 0.973 & 0.996 \\
		30 & 0.668 & 0.690 & 0.827 & 0.942 & 0.973 & 0.996 \\
		40 & 0.677 & 0.699 & 0.823 & 0.951 & 0.978 & 1.000 \\ \hline
	\end{tabular}

  \end{center}
\end{table}

\subsection{Classification accuracy}
In Experiment 2, the accuracy of the shape classification was 0.973 and the automatically selected intermediate layer was 3-1.
Table \ref{table:exp2} shows the accuracy and its rank for each layer.
\par
For the support set, the highest accuracy layer was 3-1 and its accuracy was 0.985.
This layer ranked second in accuracy for the test data, which is nearly the same as the accuracy of the support set.
Conversely, the lowest accuracy layer for the support set was 4-3, with an accuracy of 0.844.
This layer ranked 9th in the accuracy of the test data, with an accuracy of 0.881.
\par
Similarly, the ranks of the support set and the ranks of the test data are similar in most layers.
In particular, layers 2-1--3-2 rank high for both the support set and the test data.
This shows that object shape features are likely to appear in these layers.
This is likely because the lower layers are better suited to extract the object shape features.
Because the final output of the DNN in these experiments is the grasping positions, the higher layers closer to the final layer are more susceptible to the grasping positions.
This result shows that the proposed method can automatically select the intermediate layer suitable for shape classification.
\par

\begin{table}[tb]
  \begin{center}
	\label{table:exp2}
	\begin{tabular}{c|c|c|c|c}

		\multicolumn{5}{c}{TABLE II: Results of Experiment 2} \\

		\hline
		Intermediate & Support set & Test data & Support set & Test data \\
		layer & accuracy & accuracy & rank & rank \\ \hline
		1-1 & 0.948 & 0.947 & 4 & 7 \\
		1-2 & 0.948 & 0.951 & 4 & 6 \\ \hline
		2-1 & 0.963 & 0.965 & 3 & 4 \\
		2-2 & 0.970 & 0.978 & 2 & 1 \\ \hline
		3-1 & 0.985 & 0.973 & 1 & 2 \\
		3-2 & 0.948 & 0.965 & 4 & 4 \\
		3-3 & 0.933 & 0.929 & 7 & 8 \\ \hline
		4-1 & 0.919 & 0.973 & 8 & 2 \\
		4-2 & 0.911 & 0.854 & 9 & 10 \\
		4-3 & 0.844 & 0.881 & 10 & 9 \\ \hline
	\end{tabular}

  \end{center}
\end{table}

In Experiment 3, 775 grasping positions were detected for the 200 test data images.
The correct classification was achieved for 745 out of 775 grasping positions, and the accuracy was 0.961.
Conversely, the accuracy was 0.234 when not using our method.
We can see therefore that the proposed method is effective for input images that have multiple objects of the same shape.
\par
In Experiment 4, 203 grasping positions were detected for the 50 test data images.
The correct classification was achieved for 200 out of 203 grasping positions, and the accuracy was 0.985.
However, the accuracy was 0.197 when not using our method.
We can see that the proposed method is therefore also effective for input images that have multiple objects of different shapes.
\par
In Experiment 5, 187 grasping positions were detected for the 50 test data images. 
The correct classification was achieved for 184 out of 187 grasping positions, and the accuracy was 0.983.
This result shows that our framework can be used without retraining the SSD when a target object is changed.
\par
The accuracies of Experiments 2-5 are shown in Table \ref{table:exp2-5}.
The accuracy is nearly the same for all experiments.
We find that the number of objects in the input image does not affect the classification accuracy.
Therefore, we conclude that our method can refine the features for each grasping position.
\par

\begin{table}[tb]
  \begin{center}
	\label{table:exp2-5}
	\begin{tabular}{c|c|c|c|c}

		\multicolumn{5}{c}{TABLE III: Comparison of experimental results} \\

		\hline
		Experiment & Grasping  & Correct & Test data & Selected \\	
		 & positions & answers & accuracy & layer \\ \hline
		2 & 226 & 220 & 0.973 & 3-1 \\
		3 & 775 & 745 & 0.961 & 3-1 \\
		4 & 203 & 200 & 0.985 & 3-1 \\
		5 & 187 & 184 & 0.983 & 3-1 \\ \hline
	\end{tabular}

  \end{center}
\end{table}

In this paper, we implemented multi-task learning to detect the grasping positions and to classify the shape of an object.
By our proposal method, we can retrain the SVMs with only a few new labeled images, when object shape of a product or a part is changed in a factory.
We don't have to retrain the SSD for a new object.
Therefore operating costs become low.
In contrast, a end-to-end model which learned multi-task is more costly, because it needs a large amount of new labeled images of a product or a part.
In terms of cost, our framework is superior to a end-to-end model of multi-task.
\par
The addition of another task is also conceivable, for example, classifying the color of an object.
Because color is not related to the grasping positions of objects, it is likely that color features do not appear in the higher layers close to the output layer.
However, it is expected that color features will remain in the lower layers closer to the input layer.
We believe that color classification is possible in the proposed framework because the intermediate layer in which the color features remain can be selected automatically.
\par
Our framework may also be applicable to multi-task problems other than object grasping by a robot arm.
Similar to the experimental setup in this paper, we believe this method can be applied to tasks that perform detection and classification simultaneously (for example, car detection and car type classification tasks).


\addtolength{\textheight}{-8cm}   

\section{CONCLUSIONS}
In this study, we proposed a multi-task learning framework that performed a shape classification task using the features of a DNN that was trained to detect the grasping position.
Specifically, we proposed a method to refine the features for each grasping position, to select features from the optimal layer of the DNN, and to evaluate them.
\par
The results showed high accuracies for the shape classification even when the input image had different shape objects. 
We confirmed that the method was able to classify the shape of an object with a small dataset.
In addition, the results showed that our framework could be used without retraining the DNN when a target object was changed.
Therefore operating costs for a picking robot in a factory can be reduce. 
\par
Moving forward, we are experimenting with classifications other than the object shape and will extend our framework and apply it to other multi-task problems.





\end{document}